\documentclass[letterpaper, 10 pt, conference]{ieeeconf}
\usepackage{graphicx}
\usepackage{amssymb}
\usepackage{epstopdf}
\usepackage{multicol}
\usepackage{multirow}
\usepackage{fancyhdr}
\usepackage{array}
    \newcolumntype{P}[1]{>{\centering\arraybackslash}p{#1}}
    \newcolumntype{M}[1]{>{\centering\arraybackslash}m{#1}}
    
\IEEEoverridecommandlockouts 
\usepackage{amsmath}
\usepackage{mathtools}
\usepackage{tikz}
\usepackage{pdflscape}
\usepackage[acronym]{glossaries}
\usepackage{algorithm}
\usepackage{algpseudocode}
\usepackage[left=1in, right=1in]{geometry}                
\DeclarePairedDelimiterX{\norm}[1]{\lVert}{\rVert}{#1}

\renewcommand{\vec}{\mathbf}
\newcommand*{\vertbar}{\rule[-1ex]{0.5pt}{2.5ex}}
\let\oldref\ref
\renewcommand{\ref}[1]{(\oldref{#1})}

\def \encoder {\Phi}
\def \obs {\phi}
\def \decoder {\Psi}
\def \koop {K}

\setacronymstyle{long-short}
\newacronym{admd}{aDMD}{Autoencoder Dynamic Mode Decomposition}
\newacronym{dmd}{DMD}{Dynamic Mode Decomposition}
\newacronym{edmd}{eDMD}{Extended Dynamic Mode Decomposition}
\newacronym{lfd}{LfD}{Learning from Demonstration}
\newacronym{dmp}{DMP}{Dynamic Movement Primitives}
\newacronym{rl}{RL}{Reinforcement Learning}
\newacronym{lasa}{LASA}{Learning Algorithms and Systems Laboratory}

\title{\LARGE \bf
Deep Learning for Koopman--based Dynamic Movement Primitives
}

\author{Tyler Han$^{1}$ and Carl Glen Henshaw, PhD$^{2}$
\thanks{$^{1}$Tyler Han is an undergraduate intern at the Naval Center for Space Technology, U.S. Naval Research Laboratory, Washington DC, USA 20375 
        {\tt\small tyler.han@nrl.navy.mil}}%
\thanks{$^{2}$Glen Henshaw is  with the Robotics and Machine Learning Section, Naval Center for Space Technology, U.S. Naval Research Laboratory, Washington DC, USA 20375
        {\tt\small glen.henshaw@nrl.navy.mil}}%
}

\usepackage{xspace}  
\usepackage{xcolor}

\fancypagestyle{titlepage}{
	
	\fancyhead{}
	\fancyfoot{}
	\fancyfoot[CO,CE]{Distribution Statement ``A'': Approved for Public Release, Distribution Unlimited.}
	}

\begin{document}
\maketitle
\thispagestyle{titlepage}

\section*{Abstract}

The challenge of teaching robots to perform dexterous manipulation, dynamic locomotion, or whole--body manipulation from a small number of demonstrations is an important research field that has attracted interest from across the robotics community. In this work, we propose a novel approach by joining the theories of Koopman Operators and Dynamic Movement Primitives to Learning from Demonstration.  Our approach, named \gls{admd}, projects nonlinear dynamical systems into linear latent spaces such that a solution reproduces the desired complex motion.
Use of an autoencoder in our approach enables generalizability and scalability, while the constraint to a linear system attains interpretability.
Our results are comparable to the Extended Dynamic Mode Decomposition on the LASA Handwriting dataset but with training on only a small fractions of the letters.

\section{Introduction}
One facet of robotic learning entails training a robot to perform a task with a small number of demonstrations provided from an expert. Simple \gls{lfd} problems may be solved by the straightforward application of function approximation, \textit{eg} trajectory examples provided by an operator can be approximated using splines and replayed as needed. However, this approach is suitable only for problems where the problem is highly constrained: with the robot working in the same area of the workspace, identical objects being manipulated across trials, and so on. As such, this type of technique does not readily allow changes in the task such as novel trajectories or variations in timing. Such techniques have found widespread use in factory automation, where these constraints can be enforced, but are less satisfactory in unstructured environments such as agriculture, disaster response, or domestic service.

\gls{rl} is an obvious technique to consider for \gls{lfd} problems. Typically,  \gls{rl} requires long training times with many examples but an appropriate design of the architecture, along with the policy, loss, and/or reward, can result in significantly improved data efficiency\cite{ng1999policy, laud2004theory, amodei2016concrete}. Notable early work in \gls{rl} for robotic motor learning was done by Schaal \cite{schaal1997learning}. 

An alternative approach entails treating motions as an output of an underlying dynamical system. This approach is known \gls{dmp} \cite{schaal2005learning,pastor2009learning,kober2012reinforcement,ijspeert2013dynamical} and provides substantial motivation for the framing of our work. \gls{dmp}s formulate the motion generation for an \gls{lfd} problem as a basin attractor system, normally designed by hand, and usually augmented with a learned forcing term. DMPs have been successfully used for a variety of \gls{lfd} problems \cite{matsubara2011learning,davchev2020residual}. A recent survey of the various mathematical formulations and adaptations can be found in \cite{saveriano2021dmp}.

In the field of dynamical systems, a burgeoning area of study is the data-driven identification of Koopman operators\cite{kutz2016dynamic, brunton2016sindy, lusch2018deep}. As opposed to typical differential equations descriptions of a dynamical system, which formulate the system as the evolution of a finite--length state variable, Koopman theory formulates the dynamical system as the evolution of \textit{observables}. The set of observables is the set of all scalar functions on the state, and hence the Koopman operator $\mathcal{K}$ is an infinite dimensional operator. However, importantly, the Koopman operator is a \textit{linear} operator. With classic differential equations we accept nonlinearity in order to work with finite state vectors but with Koopman theory we work with a potentially infinite observation vector in order to have linear dynamics.

Although infinite--dimensional vectors are difficult to compute with, Koopman analysis suggests that in some cases, a system of interest can be either exactly or approximately represented with finite approximations to the Koopman operator \cite{kutz2016dynamic}. Obviously, infinite--dimensional operators are difficult to store and compute but Koopman analysis suggests that in some cases, a system of interest can be either exactly or approximately  represented with finite approximations to the Koopman operator \cite{kutz2016dynamic}. This involves identifying a particular set of functions known as \textit{observables} on the state of the system.

Classically, there are basis sets of functions that are known to be useful in the approximation of the Koopman operator; these include truncated monomial and polynomial expansions; transcendental functions; delay functions; and radial basis functions. However, the identification of appropriate observable functions for specific applications remains one of the fundamental challenges in Koopman theory. As a consequence, deep learning frameworks have attracted interest for approximating observable functions and the resultant latent space representations they induce \cite{lusch2018deep}.

Here, we join the the theories of \glspl{dmp} and Koopman operators in a novel approach to robotic motion. As a preliminary experiment, we verify the approach using a handwriting dataset compiled by the \gls{lasa}. We refer to our approach as \gls{admd} in reference to the well--studied techniques \gls{dmd} and \gls{edmd}.

\subsection*{Related Work}

Notably, Lian and Jones \cite{lian2019learning} provide a rigorous framework for learning both the observation functions and the Koopman operator from data. They used Gaussian processes, a universal approximator, to learn observation functions and the resultant latent space, and demonstrated good results on the \gls{lasa} dataset \cite{lian2019learning}. Lian and Jones make no claim that the learned observation functions are suitable for trajectories or tasks that are not in the training set, however, and demonstrated performance on individual characters. As a consequence, a disadvantage of this approach is the need to train a separate dynamical model for each character individually, including the observation functions, and retrain if new character strokes are desired. Here, we demonstrate equivalent performance to Lian and Jones with character stroke models using a fraction of the character strokes as training data, without requiring retraining for new characters.

\section{Background}
We assume that there exists an underlying dynamical system which dictates the flow of the states $x\in S$ of a system. Regardless of what information is realistically obtainable by the designer, $x$ abstractly contains all the information needed to describe the system's instantaneous state. In the context of discrete systems, we denote these evolutions $x_{i+1} = f(x_i)$ where $f:S\rightarrow S$ represents the underlying dynamics, which in general are nonlinear. 

As mentioned, Koopman theory formulates the dynamical system as the evolution of observables, which are scalar-valued functions of the state. This is typically described via operator theory as
\begin{eqnarray}
&& \mathcal{K} g = g \circ f \\
\Rightarrow && \mathcal{K} g(x_{i}) = g(f(x_{i})) = g(x_{i+1})
\label{eqn:koopman}
\end{eqnarray}
where $g:S\rightarrow \mathbb{R}$ is an observable function.
 A comprehensive overview of Koopman operator theory is provided in the book by Kutz \textit{et al} \cite{kutz2016dynamic}.

Dynamic Mode Decomposition (\gls{dmd}) is one technique for determining a finite approximation to the Koopman operator. In \gls{dmd}, data $X$, representing the state vector over time, is captured from the real system and the dynamics are approximated using a least-squares solution. Let $\vec x_i$ be the columns of $X\in \mathbb{R}^{n \times m}$. Denote 
$X_1=\begin{bmatrix} 
	\vec x_1 &\hdots &\vec x_{n-1}
\end{bmatrix}$ and 
$X_2=\begin{bmatrix} 
	\vec x_2 &\hdots &\vec x_{n}.
\end{bmatrix}$
By assuming that the system is linear, one can solve the equation $X_2 \approx \tilde{A}X_1$ for $\tilde{A}$ via a pseudoinverse. This solution represents a least--squares fit to a linear dynamical system. \gls{edmd} generalizes this technique to the space of observables such that $Y_2\approx\tilde A_{e}Y_1$ where $Y_1=g(X_1)$, $Y_2=g(X_2)$, and $g:\mathbb{R}^{n \times m} \rightarrow \mathbb{R}^{k \times m}$ represents a set of $k$ observable functions $g_i:\mathbb{R}^{n}\rightarrow \mathbb{R}$ performed on each column of the input. Choosing an appropriate set of observable functions is in general a difficult problem. Approximating these functions using neural networks is one of the goals of this work.


\section{Approach}

\subsection{Autoencoder Dynamic Mode Decomposition (aDMD)}

\begin{figure}[t]
	\centering
	\includegraphics[width=2.8in]{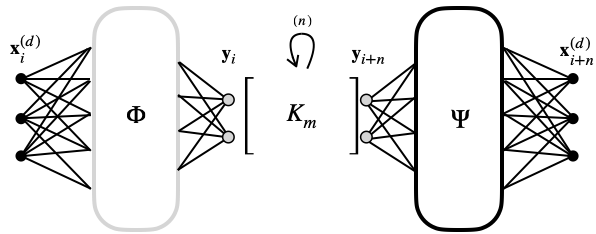}
	\caption{Diagram of \gls{admd} Architecture, composed of an encoder $\encoder$ and decoder $\decoder$ surrounding an eDMD-computed recurrent state transition matrix, $\koop_{m}$.}
	\label{dkfig}
\end{figure}

Our model is largely adapted from \cite{lusch2018deep}, where the authors employ an autoencoder but use an auxiliary network in the identification of a Koopman operator. In contrast, our approach is to focus solely on the discovery of a set of observable functions for representing multiple trajectories as a discrete spectrum Koopman operator (see Fig. \ref{dkfig}). We further propose that the set of identified observables (encoder) is extensible to other ``similar'' trajectories not specifically trained on.

The process of aDMD is equivalent to eDMD by using latent representations from the autoencoder as the set of observables as a linear system. Equation \ref{eqn:koopman} imposes requirements on the latent representation so that propagating a trajectory from an initial condition in latent space and using the decoder to return to delay space, we can compute the linear, prediction, and reconstruction losses. Linear loss \ref{linearloss} ensures that the latent space of the trajectory is indeed linear, as dictated by a set of linearizing observable functions. Prediction loss \ref{predloss} ensures returning to delay space from latent space corresponds to the correct points in the trajectory. Reconstruction loss \ref{reconsloss} is the standard loss as dictated by an autoencoder, requiring that data transformed by the encoder can be recovered with maximal accuracy. Finally, a regularization term is added to prevent overfitting.

Formally, let $\encoder:\mathbb{R}^{n_d}\rightarrow \mathbb{R}^m$ and $\decoder:\mathbb{R}^m\rightarrow \mathbb{R}^{n_d}$ represent the encoder and decoder networks of the autoencoder, respectively. Let $\koop:\mathbb{R}^m\rightarrow \mathbb{R}^m$ represent the approximated Koopman operator, which is calculated using
$\encoder(\cdot)=\begin{bmatrix}\obs_1(\cdot) & \obs_2(\cdot) & \hdots & \obs_m(\cdot)\end{bmatrix}^T$
where each $\obs_j:\mathbb{R}^{n_d}\rightarrow \mathbb{R}$ is an eDMD observable function. $\koop_{m}$ is calculated via a pseudoinverse,
\begin{equation}
	\koop_{m} = Y_2Y_1^\dagger
	\label{dmdeq}
\end{equation}
where the latent state data matrices $Y_1$ and $Y_2$ are transformed snapshots of the data as in
\begin{align}
	Y_1  = &
	\begin{bmatrix} 
		\vertbar & \vertbar &  & \vertbar \\ 
		\encoder(\vec x^{(d)}_1) & \encoder(\vec x^{(d)}_2) & \hdots & \encoder(\vec x^{(d)}_{N-1}) \\
		\vertbar & \vertbar &  & \vertbar \\ 
	\end{bmatrix} \qquad \\
	Y_2 = &
	\begin{bmatrix} 
		\vertbar & \vertbar &  & \vertbar \\ 
		\encoder(\vec x^{(d)}_2) & \encoder(\vec x^{(d)}_3) & \hdots & \encoder(\vec x^{(d)}_{N}) \\
		\vertbar & \vertbar &  & \vertbar \\ 
	\end{bmatrix}.
	\label{shiftedtransformed}
\end{align}

For a single trajectory, we use the loss function
\begin{align}
\label{singleloss}
\mathcal{L} &= \mathcal{L}_{lin} + \alpha\mathcal{L}_{pred} + \mathcal{L}_{recon} + \beta\norm{\mathbf{W}}^2_2 \\
\label{linearloss}
\mathcal{L}_{lin} &= \sum_{m=1}^N \norm{\koop^m\encoder(\vec x_1) - \encoder(\vec x_{m+1}) }_{\textrm{\tiny MSE}} \\
\label{predloss}
\mathcal{L}_{pred} &= \sum_{m=1}^N \norm{\decoder(\koop^m\encoder(\vec x_1)) - \vec x_{m+1} }_{\textrm{\tiny MSE}} \\
\label{reconsloss}
\mathcal{L}_{recon} &= \sum_{i=1}^N \norm{\decoder(\encoder(\vec x_i)) - \vec x_{i} }_{\textrm{\tiny MSE}}
\end{align}
where $\alpha$ and $\beta$ are hyperparameters. Training over multiple trajectories, we represent the final loss function as the sum of the losses for each individual trajectories.

\subsection{Delay Coordinates}

\begin{figure}[t]
	\centering
	\includegraphics[width=2.8in]{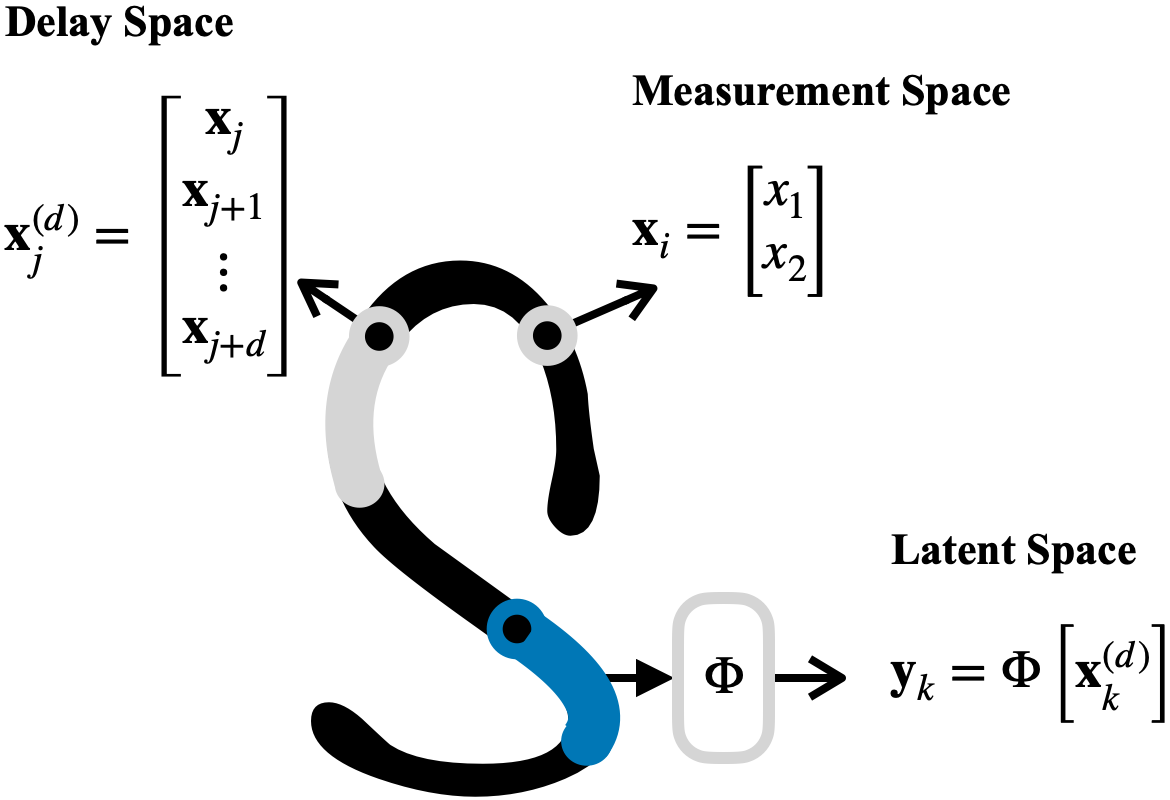}
	\caption{Visualization of delay and latent space with respect to measurement space}
	\label{space-visual}
\end{figure}

Delay coordinates were introduced into \gls{dmd} for discovering dynamics with standing waves \cite{kutz2016dynamic} and, recently, linearity and forcing in strongly nonlinear (chaotic) dynamical systems \cite{brunton2016havok}. They are a simple yet powerful way to compensate for data-sparse problems and allows the model to learn without overfitting the task space.
In our case, the employment of delay coordinates can be seen as a way to ``chunk'' segments of a trajectory over time (Fig. \ref{space-visual}). 

A coordinate $\vec x_i$ at time $i$ is delayed once if instead of $\vec x_i$ we collect $\vec x_{i+1}$ at time $i$. We say that our data is augmented with $d$ delay coordinates when each datapoint is delayed $d$ times, each time augmenting the state vector with the resulting delayed state:
\begin{equation}
	X^{(d)} = 
	\begin{bmatrix} 
		\vec x_1 & \vec x_2 & \hdots & \vec x_{N-d} \\		
		\vec x_2 & \vec x_3 & \hdots & \vec x_{N-d+1} \\		
		\vdots & \vdots & \ddots & \vdots \\
		\vec x_{d+1} & \vec x_{d+2} & \hdots & \vec x_{N} \\		
	\end{bmatrix}
\end{equation}
For brevity, we will henceforth refer to the $i$'th column of the delayed data $X^{(d)}$ as $\vec x^{(d)}_i = \begin{bmatrix}\vec x_i^T & \vec x_{i+1}^T & \hdots & \vec x_{i+d}^T\end{bmatrix}^T$. Note that the dimensionality of this vector is $n_d = n(d+1)$ and the duration (number of columns) of the delayed trajectory is now $N_d = N-d$.

\subsection{Eigenmode Filtering}
Even when using truncated SVD pseudoinverses, \gls{admd} occasionally finds an ``unstable'' eigenmode --- an eigenvalue of $\koop$ greater than 1. We can filter out this eigenmode by subtracting it in latent space before returning to measurement space by determining the indexing set $F$ whose corresponding eigenmodes are not unstable,
\begin{equation}
	\tilde{X} = \decoder\left[\sum^m_{j=1}P^{(k)}_j - \sum_i P^{(k)}_i\right]
	= \decoder\left(\sum_{f\in F} P^{(k)}_f\right)
	\label{modefiltereq}
\end{equation}

\subsection{Summary}
In summary of our algorithm, we  propagate the initial coordinate using the corresponding Koopman system: 
\begin{enumerate}
	\item Given the data $X$ for a trajectory, obtain $X^{(d)}$ for the delay $d$ with which the training data was augmented. 
	\item Identify $\koop_{m}$ using equations \ref{dmdeq}, \ref{shiftedtransformed}, and \ref{modefiltereq}
	\item Let $\vec x_1^{(d)}$ be the first delay coordinate datapoint in the reconstruction
	\item Other points in the reconstruction in delay coordinates are obtained using the rule 
	\[\vec x_{i}^{(d)} \approx \decoder(\koop_{m}^{i-1}\cdot\encoder(\vec x_1^{(d)}))\]
	\item The full reconstruction in measurement space is given by the first $n$ rows of the matrix whose $i$th column for $i\in[1,2,\hdots,N_d]$ is the reconstructed datapoint $\vec x_i^{(d)}$
\end{enumerate}


\section{Results}

\begin{table*}[t]
\begin{center} 
\begin{tabular}{ c p{2.5cm} p{3.5cm} p{3.5cm} }
	& \gls{admd} & Polynomial eDMD ($n=3$)& Polynomial \gls{edmd} ($n=4$) \vspace{0.05in}\\
	\multicolumn{4}{c}{Average Error per Trajectory (Single Reconstruction)} \\
	\hline
	Prediction Error  & 0.0242399 & 0.1961263 & 0.0292222 \\
	Linear Error  & 0.0681664 & 1360.897 & 5407.519  \vspace{0.02in}\\
	\multicolumn{4}{c}{Average Error per Trajectory (Noisy Reconstruction)} \\
	\hline
	Prediction Error  & 0.0561916 & 0.5516364 & 0.0705166 \\
	Linear Error  & 684.5577 & 4082.349 & 7291.988 \\
\end{tabular}
\caption{Comparison of errors for \gls{admd} and polynomial \gls{edmd}. }
\label{errortable}
\end{center}
\end{table*}

\begin{table*}[t]
\renewcommand{\arraystretch}{4}
\centering
\begin{tabular}{ m{0.75in} m{2.2in} m{2.2in} }

& Single Reconstruction & Noisy Reconstructions \\ 
\gls{admd}
&\includegraphics[width=2.1in]{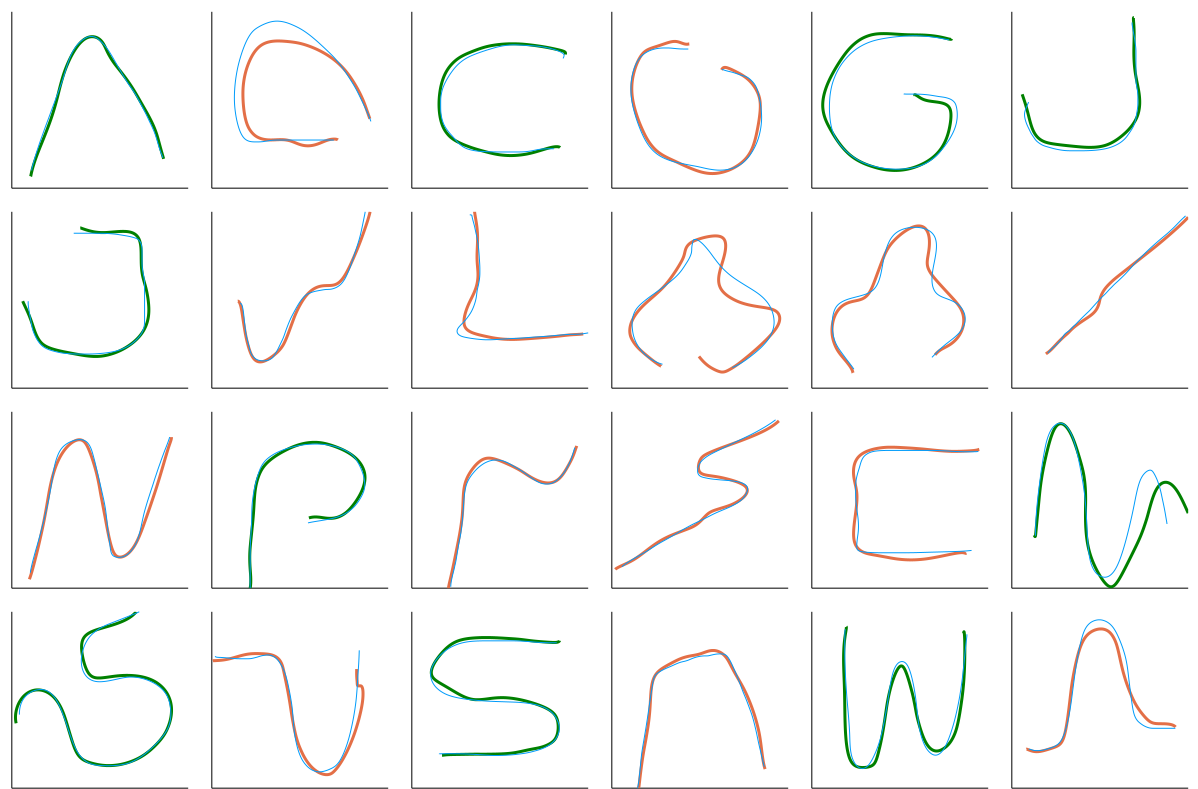}
&\includegraphics[width=2.1in]{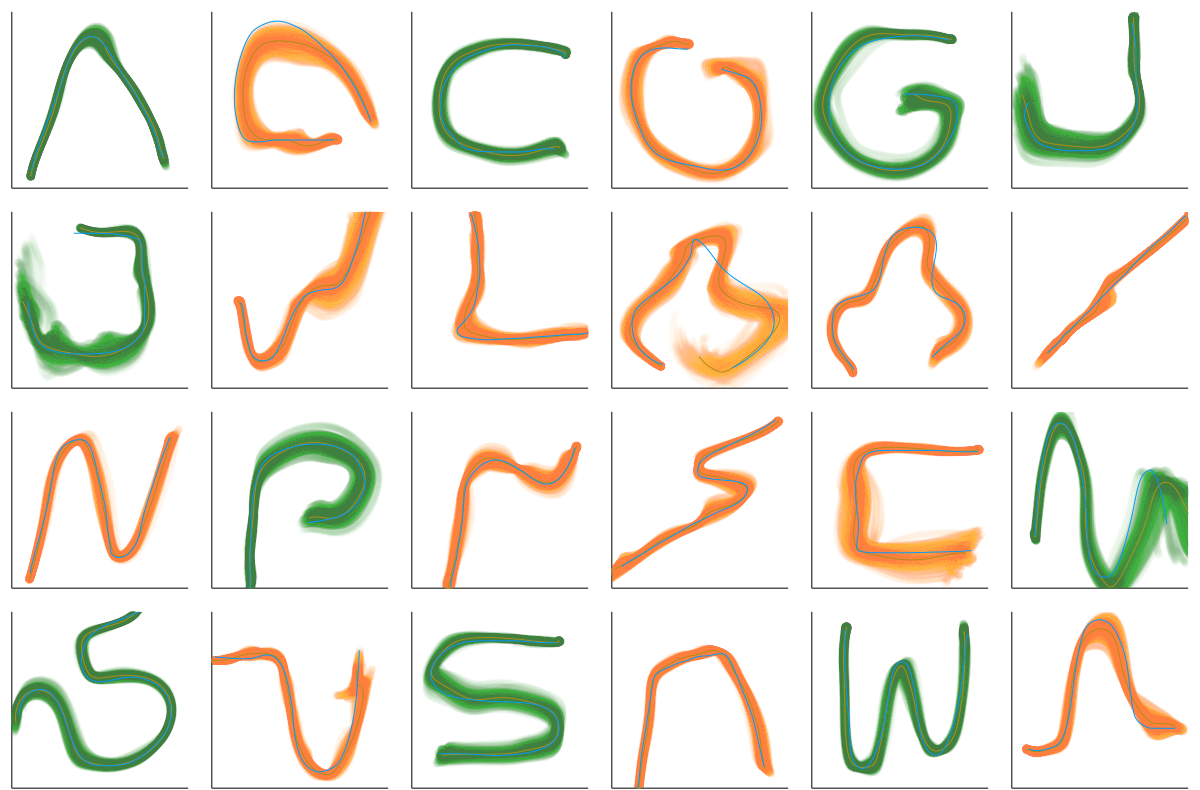} \\
pDMD ($n=3$) 
&\includegraphics[width=2.1in]{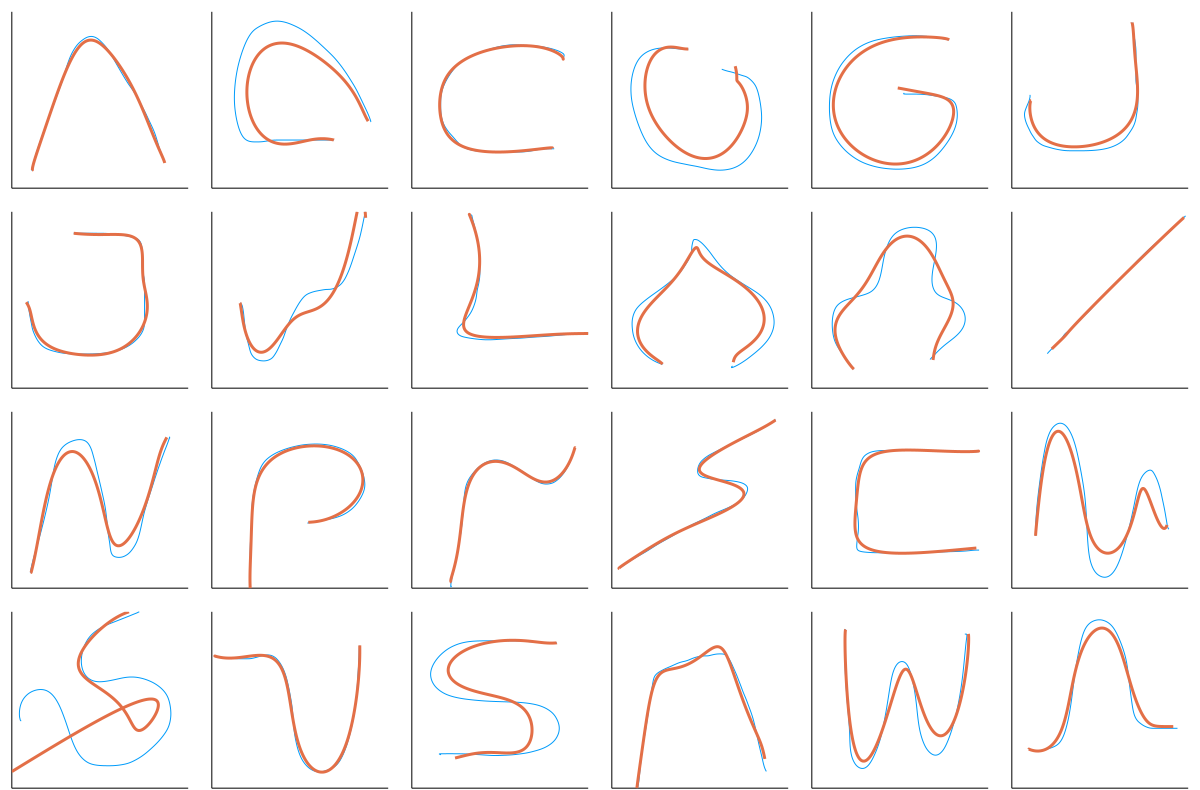}
&\includegraphics[width=2.1in]{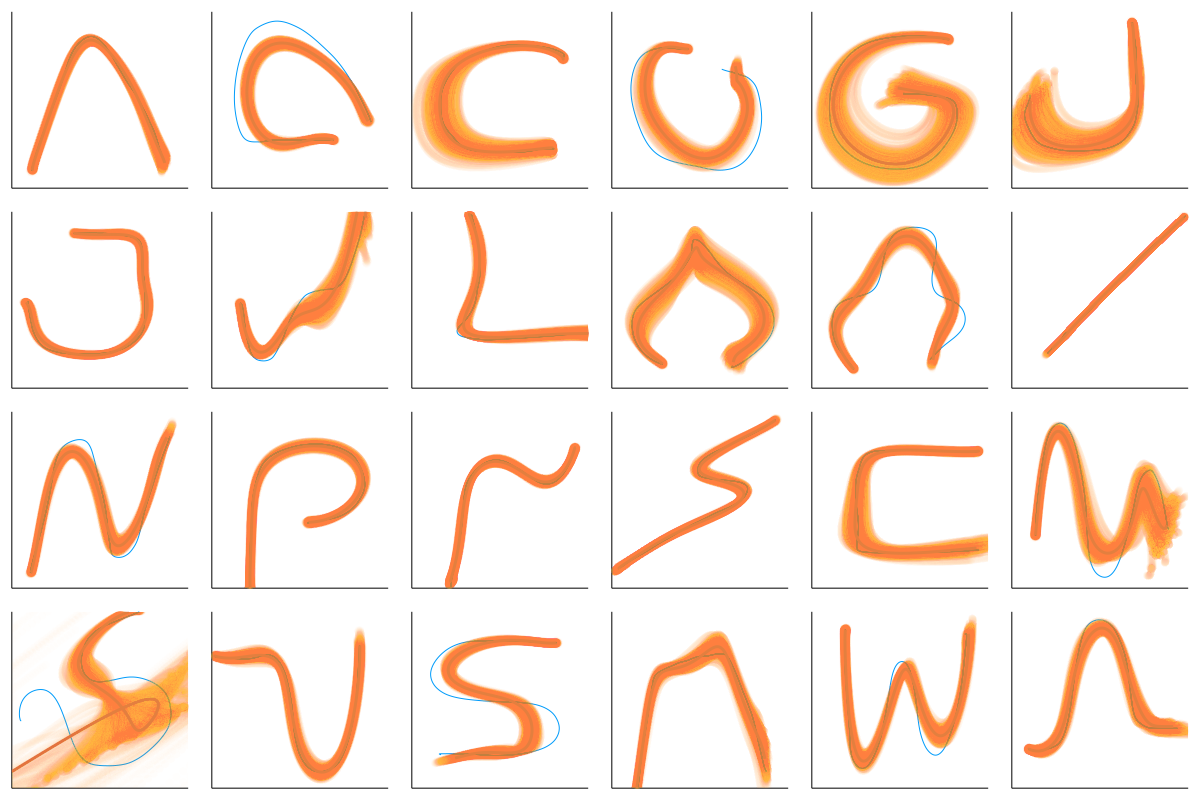} \\
pDMD ($n=4$) 
&\includegraphics[width=2.1in]{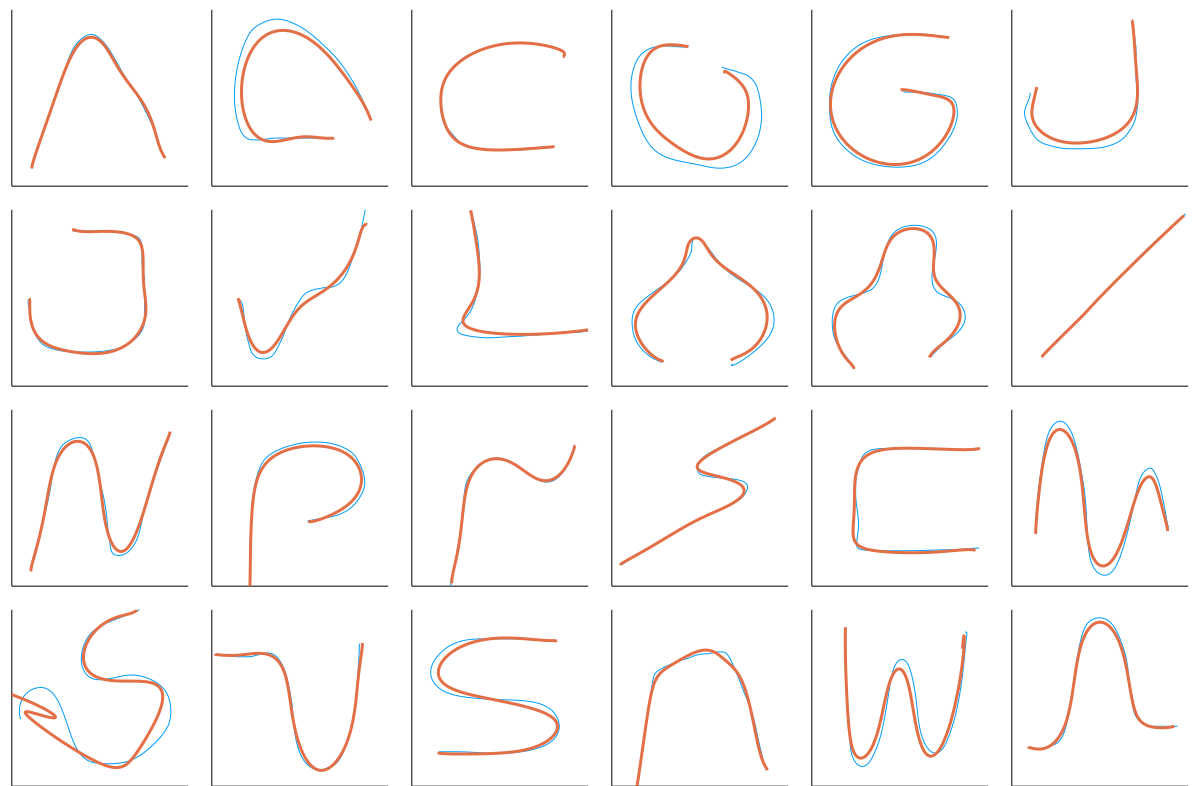}
&\includegraphics[width=2.1in]{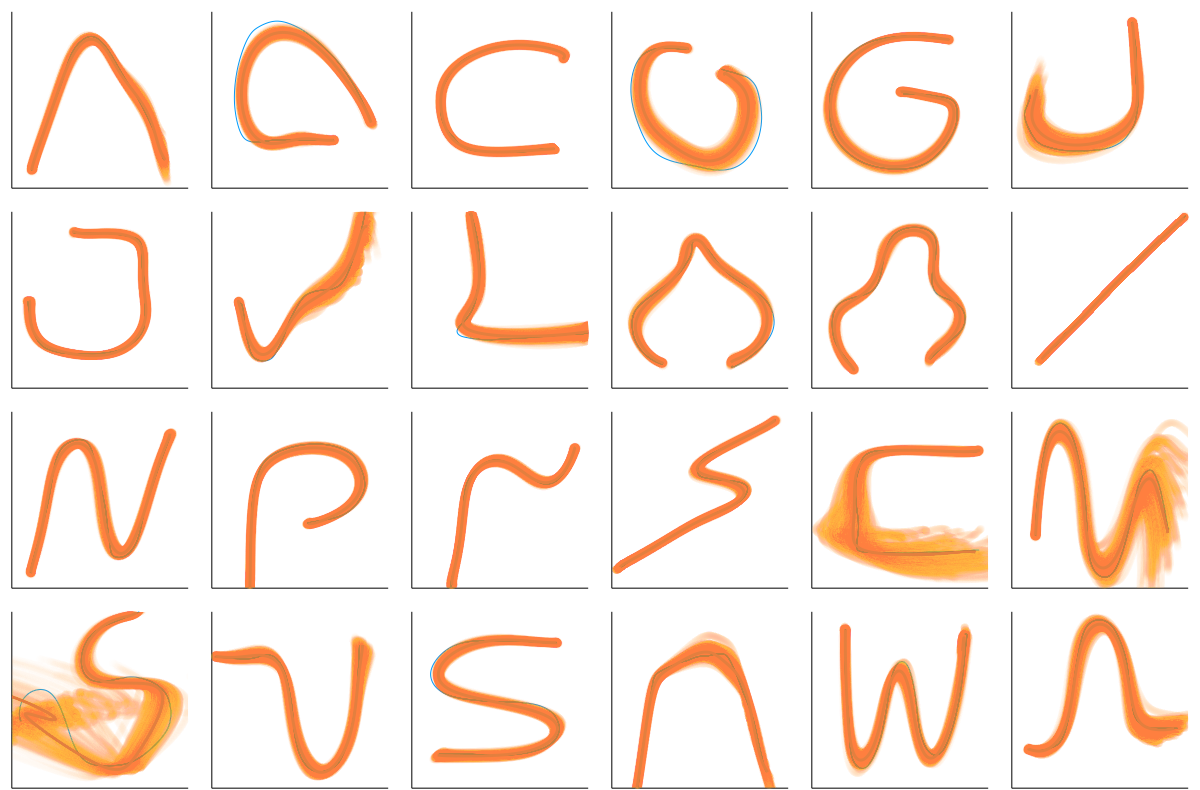} \\

\end{tabular}
\caption{Reconstruction plots comparing \gls{admd} and \gls{edmd}. Original characters are in blue. Reconstructed training set characters are in green and reconstructed test set examples in orange.
 }
\label{comparisontable}
\end{table*}

We validate our approach on the \gls{lasa} Handwriting dataset \cite{khansari2011learning}, which contains 26 handwritten, single-stroke ``characters". Data augmentation was used to improve the robustness and generalizability of the \gls{admd}.  Augmenting the training set with random noise expands the training region and yields Koopman systems which can recreate letters even under perturbations of the initial conditions. This practice is also sometimes referred to as ``motor babbling'' \cite{saegusa2009active}.

We evaluate the average prediction and linear error for single reconstructions, which take the datapoint $\vec x_1$ as the first point in the reconstruction, and the average prediction and linear error for noisy reconstructions, which take $\tilde{\vec x}_i \sim \mathcal{N}(\vec x_i,0.05^2)$. The exact error equations are provided in the appendix.

Results are shown in Tables \ref{errortable} and \ref{comparisontable}. \gls{admd} reconstruction error is superior to 3--rd and 4--th order polynomial \gls{edmd}. Notable is \gls{admd}'s inherently compressive architecture. The number of effective states acted on by the autoencoder (delay space) is two times as large as the state in the learned linear system (latent space). Yet, the results are superior to polynomial \gls{edmd}.  This provides evidence that the autoencoder achieves some level of compression perhaps adaptable to high--dimensional systems. In comparison, classical observable selection or dictionary methods (such as polynomial \gls{edmd}) typically explode or become intractable in high-dimensional systems \cite{williams2014kdmd}.


\section{Discussion \& Further Work}

We have shown that every example in the LASA Handwriting set can be described by observables identified by \gls{admd}. The  autoencoder $\encoder(\cdot)$ needs only one training example per trajectory and generalizes to the space of characters not yet seen. Test set characters are fit robustly, generally withstanding the same magnitude of perturbations on which the observables were trained. 

We justify the framing of this approach as inspired by \gls{dmp}s because, like \gls{dmp}s, \gls{admd} formulates the motion generation law for an \gls{lfd} task as a dynamical system. Unlike other \gls{dmp} approaches, however, \gls{admd} allows us to automatically identify the appropriate dynamics directly from data, instead of hand--designing an attractor basin as is usually the case with \gls{dmp}s. 

For future work we intend to test on humanoid and quadruped locomotion and whole--body manipulation data, with the hypothesis that the movements of particular interest in those systems will exhibit analogous latent spaces, ultimately compressing and linearizing the dynamics and controls of interest to engineers. We also intend to investigate applying linear control theory within the linear latent space, with the intention of robustly handling system perturbations and external disturbances.



\section{Appendix}
\subsection{Results Error Calculation}
For a given trajectory's data $\vec x_i\in \mathbb{R}^n$, errors for single reconstructions were calculated using
\begin{align}
	\mathcal{E}_{lin} &= \sum_{m=1}^N \norm{\koop^m g(\vec x_1) - g(\vec x_{m+1}) }_{\textrm{\tiny MSE}} \\
	\mathcal{E}_{pred} &= \sum_{m=1}^N \norm{h(\koop^m g(\vec x_1)) - \vec x_{m+1} }_{\textrm{\tiny MSE}} 
\end{align}
where $g(\cdot)$ is the set of observables for the method and $h(\cdot)$ is the mapping back to measurement space (which includes taking the first $n$ rows for \gls{admd}).

For noisy reconstructions, 100 initial conditions were perturbed with a random variable sampled from a truncated multivariate normal distribution $\vec{\tilde{x}}_i \sim \mathcal{N}(\vec x_i,0.05^2)$ and propagated into a trajectory.
\def \noisyx {\vec{\tilde{x}}}
\begin{align}
	\mathcal{E}_{n\; lin} &= \frac{1}{100}\sum^{100}_{i=1}\sum_{m=1}^N \norm{\koop^m g(\noisyx_1) - g(\vec x_{m+1}) }_{\textrm{\tiny MSE}}
	\label{noisylin} \\
	\mathcal{E}_{n\; pred} &=\frac{1}{100}\sum^{100}_{i=1} \sum_{m=1}^N \norm{h(\koop^m g(\noisyx_1)) - \vec x_{m+1} }_{\textrm{\tiny MSE}} 
	\label{noisypred}
\end{align}

\subsection{Training Details}

When training against noise, $\vec d_i \sim \mathcal{N}(\vec 0,0.05^2)$ was added to every state $\vec x^{(d)}_i$ in every trajectory.

Hyperparameter values used in the results of this paper are: $\alpha=100, \beta=10^{-12}$ are used to tune the prediction and regularization losses, respectively. $m=20$ is the dimension of the Koopman operator. $d=20$ is the number of augmented delay coordinates. $f$, the activation function in the autoencoder hidden layers, was the standard ELU function; hid\_width=20 and num\_hidden=2 are the width and number of the autoencoder hidden layers, respectively. The model was trained using the Adam optimizer as implemented in the Julia programming language and Flux.jl library \cite{Flux.jl-2018}.

\clearpage
\bibliographystyle{ieeetr}
\bibliography{ref}

\end{document}